\documentclass[journal]{IEEEtran}

\usepackage[letterpaper,top=2cm,bottom=2cm,left=2cm,right=2cm,marginparwidth=1.25cm]{geometry}

\usepackage{tikz}
\usepackage{float}

\usetikzlibrary{decorations.fractals,spy}
\usepackage{graphicx}
\usepackage{amsmath}
\usepackage{amssymb}
\usepackage{booktabs}
\usepackage{bm}
\usepackage{algorithmic}
\usepackage{algorithm}
\usepackage{upgreek}
\usepackage[colorlinks=true, allcolors=blue]{hyperref}
\usepackage{multirow}
\usepackage{graphicx}
\usepackage{threeparttable}
\usepackage[caption=false]{subfig}
\usepackage[normalem]{ulem}
\useunder{\uline}{\ul}{}

\newcommand{\fig}[1]{Fig.~\ref{#1}}
\newcommand{\beq}{\begin{equation}}
	\newcommand{\eeq}{\end{equation}}
\newcommand{\bqa}{\begin{eqnarray}}
	\newcommand{\eqa}{\end{eqnarray}}

\graphicspath{ {./figures/} }
\title{QiNN-QJ: A Quantum-inspired Neural Network with Quantum Jump for Multimodal Sentiment Analysis}

\author{Yiwei~Chen,~\IEEEmembership{Member,~IEEE,}
        Kehuan~Yan, 
        Yu~Pan,~\IEEEmembership{Senior~Member,~IEEE,}
        Daoyi~Dong,~\IEEEmembership{Fellow,~IEEE}
\thanks{This work was supported in part by the National Natural Science Foundation of China under Grant 62503415, and in part by the Australian Research Council’s Future Fellowship Funding Scheme under Project FT220100656. (\textit{Yiwei Chen and Kehuan Yan contributed equally to this work.}) (\textit{Corresponding author: Yiwei Chen.})}
\thanks{Y. Chen is with the School of Engineering, Yunnan University, Kunming, 650500, China. (email: yiweichen@ynu.edu.cn).}
\thanks{K. Yan is with the College of Computer and Data Science, Fuzhou University, Fuzhou, 350100, China.}
\thanks{Y. Pan is with the Institute of Cyber-Systems and Control, College of Control Science and Engineering, Zhejiang University, Hangzhou, 310027, China. (email: ypan@zju.edu.cn).}
\thanks{D. Dong is with the Australian Artificial Intelligence Institute, Faculty of
Engineering and Information Technology, University of Technology Sydney,
Sydney, NSW 2007, Australia (e-mail: daoyidong@gmail.com).}
}

\begin{document}
\maketitle

\begin{abstract}
Quantum theory provides non-classical principles, such as superposition and entanglement, that inspires promising paradigms in machine learning. However, most existing quantum-inspired fusion models rely solely on unitary or unitary-like transformations to generate quantum entanglement. While theoretically expressive, such approaches often suffer from training instability and limited generalizability. In this work, we propose a Quantum-inspired Neural Network with Quantum Jump (QiNN-QJ) for multimodal entanglement modelling. Each modality is firstly encoded as a quantum pure state, after which a differentiable module simulating the QJ operator transforms the separable product state into the entangled representation. By jointly learning Hamiltonian and Lindblad operators, QiNN-QJ generates controllable cross-modal entanglement among modalities with dissipative dynamics, where structured stochasticity and steady-state attractor properties serve to stabilize training and constrain entanglement shaping. The resulting entangled states are projected onto trainable measurement vectors to produce predictions. In addition to achieving superior performance over the state-of-the-art models on benchmark datasets, including CMU-MOSI, CMU-MOSEI, and CH-SIMS, QiNN-QJ facilitates enhanced post-hoc interpretability through von-Neumann entanglement entropy. This work establishes a principled framework for entangled multimodal fusion and paves the way for quantum-inspired approaches in modelling complex cross-modal correlations.
\end{abstract}


\section{Introduction}
\IEEEPARstart{M}{ultimodal} sentiment analysis aims to infer human emotions by integrating heterogeneous data sources such as text, audio, and visual signals, thereby capturing the multifaceted nature of human communication \cite{das2023multimodal}. Data-driven fusion methods, including tensor-based neural networks \cite{zadeh2017tensorFusion} and attention mechanisms \cite{zhang2023learningAdaptive}, have significantly improved predictive accuracy over traditional approaches \cite{soleymani2017survey}. However, recent studies suggest that human cognition in sentiment tasks often deviates from classical probability assumptions \cite{pothos2022quantumcognition}, posing challenges for conventional fusion frameworks. These models struggle to capture inseparable intermodal dependencies and context-sensitive correlations, which limits their expressive power. Moreover, they lack interpretability both in terms of model transparency and post-hoc explainability \cite{Lipton2018TheMO}, which is critical for understanding modality contributions and building trust in decision-making.

Quantum information and computing have emerged as transformative fields, offering both practical solutions to quantum system challenges and a theoretical foundation for interdisciplinary research \cite{du2025quantum}. Particularly, classical machine learning techniques are increasingly applied to quantum domains, addressing key issues in quantum control \cite{dong2019learning}, error correction \cite{chen2016quantum}, and system identification \cite{du2024exploring}, thereby optimizing quantum system performance and efficiency \cite{dunjko2018machine}. Simultaneously, quantum machine learning algorithms, including quantum neural networks \cite{chen2022residual, zheng2024quantum}, quantum kernel methods \cite{thanasilp2024exponential, wang2025power}, and quantum reinforcement learning \cite{li2020quantum, wei2021deep}, leverage quantum principles such as superposition \cite{li2019cnmInterpretable}, entanglement \cite{2023QuantumEntanglementEmbedding}, and interference \cite{zhang2018quantum} to develop learning paradigms. These designs not only enhance model performance but also improve interpretability, providing new insights into complex data representations and reasoning. 

In quantum mechanics, entanglement refers to a joint state of a system that cannot be factored into independent subsystems, where the measurement of one part instantaneously affects the others \cite{horodecki2009quantum}. Similarly, in multimodal sentiment analysis, data from different modalities often exhibit inseparable and cross-modal correlations. For example, sarcasm detection requires a joint interpretation of text, prosody, and facial expressions, where meaning arises not from any single modality but from their collective configuration. This analogy positions quantum entanglement as a natural framework for modeling such complex interactions.

Recent studies have explored quantum-inspired methods to leverage the concept of entanglement in multimodal learning. Specifically, Quantum-inspired Multimodal Sentiment Analysis (QMSA) \cite{ZHANG201821qmsa} represents multimodal inputs as density matrices and employ quantum-like interference frameworks for decision-level fusion. The Quantum Multi-modal Fusion (QMF) method \cite{li2021quantumfusionvideo} models intra-modal semantics through superposition and cross-modal interactions through entanglement. Similarly, a quantum probability-based model \cite{liu2023quantum} addresses contextual incompatibilities beyond classical fusion methods. More recently, \cite{yan2025quantum} proposed a quantum-like fusion module that integrates Long Short-Term Memory (LSTM) with quantum entanglement to enable dynamic and context-aware multimodal fusion. However, existing quantum-inspired fusion models predominantly rely on unitary evolution to generate multimodal entanglement. While unitary transformations are theoretically expressive and globally reachable, their unrestricted capacity often leads to instability and poor generalizability in practice. Furthermore, pure unitary evolution lacks inherent mechanisms for controlling the strength of entanglement, making it challenging to achieve stable convergence when modelling noisy, incomplete or imbalanced multimodal data.


In this work, we propose a dissipative perspective for quantum-inspired neural networks, introducing a mechanism that balances expressive power with stability and generalization. Specifically, we develop a Quantum-Inspired Neural Network with Quantum Jump (QNN-QJ) for multimodal sentiment analysis. Specifically, each input modality is first embedded as a quantum pure state, represented by a unit-norm complex-valued vector that encodes the superposition of semantic interpretations. The initial joint state of the multimodal data is constructed via the tensor product of these individual modality states. To capture inseparable inter-modal dependencies, the product state is transformed into an entangled state via the QJ method \cite{plenio1998quantum}, implemented as a trainable entanglement generation module. Unlike prior works limited to unitary operators, QiNN-QJ jointly learns Hamiltonian and Lindblad parameters, integrating coherent evolution with dissipative processes. This design introduces structured stochasticity and steady-state attractor properties, thereby enabling controllable shaping of multimodal entanglement while mitigating overfitting and enhancing interpretability.
Following the entanglement embedding, quantum-like features are extracted by computing inner products between the resulting entangled state and a set of trainable quantum measurement vectors  \cite{nielsen2010quantum}. 
Importantly, quantitative entanglement measures can be directly applied to the learned representations, enabling explicit post-hoc analysis and visualization of inter-modal entanglement. 
We validate the proposed model through experiments on three widely-used benchmark datasets, including CMU-MOSI, CMU-MOSEI and CH-SIMS. QiNN-QJ demonstrates superior performance as well as improved interpretability compared to State-Of-The-Art (SOTA) baselines. 

The main contributions are summarized as follows.
\begin{itemize}
\item Dissipative Quantum-inspired Modelling: We introduce a dissipative quantum perspective into multimodal fusion by formulating a neural module grounded in the QJ method. This module enables end-to-end learning of both Hamiltonian and Lindblad operators, allowing the network to emulate open quantum system dynamics. Through this mechanism, controllable cross-modal entanglement can be effectively generated to capture complex dependencies among modalities.

\item Robust and Transparent QiNN Architecture: In QiNN-QJ, each modality is encoded as a quantum pure state representing a superposition of semantic interpretations. The QJ module transforms initially separable multimodal product states into entangled representations with rich non-classical intermodal correlations. These entangled states are then projected through learnable quantum measurement operators to yield classical probability distributions for sentiment prediction.

\item Superior Performance and Interpretability: QiNN-QJ achieves the SOTA performance on CMU-MOSI, CMU-MOSEI and CH-SIMS datasets, outperforming conventional, quantum-inspired, and large language models. In addition, it enhances interpretability by providing a transparent mechanism for multimodal fusion and supporting post-hoc analysis of entanglement strength via von Neumann entanglement entropy.

\end{itemize}

This paper is organized as follows. Section~\ref{Sec2:Pre} introduces the preliminaries and reviews related work. Section~\ref{Sec3:Method} outlines the proposed network design and presents the modules in details. Section~\ref{Sec4:Results} presents experimental results on three benchmark datasets. Section~\ref{Sec5:Analysis} discusses post-hoc interpretability and anti-interference analysis. Finally, Section~\ref{Sec6:Conclusion} provides the conclusion.

\section{Preliminaries} \label{Sec2:Pre}
\subsection{Quantum State}
\subsubsection{Pure State}
Quantum particles can exist in a superposition of multiple states simultaneously. A pure state of a quantum system is conventionally described using Dirac notation, where the \textit{ket} notation, represented as $|\psi\rangle$, corresponds to a state vector. This vector is a complex-valued column and provides a compact yet comprehensive representation of the full information of the quantum state. The \textit{bra} notation, denoted as $\langle\psi|$, is the conjugate transpose of the \textit{ket}, mathematically expressed as $\langle\psi| = (|\psi\rangle)^\dagger$. Specifically, a pure state can be defined as
\begin{equation}
\label{eq:pure_stat}
    |\psi\rangle=\sum_{i=0}^{n-1}\alpha_i|i\rangle,
\end{equation}
where the complex number \(\alpha_i\) satisfying \(\sum_{i=0}^{n-1}|\alpha_i|^2=1\) and \(|i\rangle\) is the orthonormal basis.

Similarly, a single phrase, image, or video can inherently convey multiple layers of meaning simultaneously. This characteristic aligns naturally with the use of quantum pure states for representing internal features. Specifically, the informational coherence preserved within quantum pure states plays a pivotal role in the multimodal fusion process, facilitating the seamless integration of diverse modalities. By maintaining this coherence, a more unified and comprehensive understanding of complex multimodal scenes can be achieved.

\subsubsection{Mixed State}
A mixed state represents the statistical description of an ensemble of pure states, which can be mathematically expressed using the density matrix as
\begin{equation}\label{eq:get density matrix}
    \rho =\sum_{i=1}^mp_i|\psi_i\rangle\langle\psi_i|,
\end{equation}
where \(|\psi_i\rangle\) denotes the quantum pure state of the subsystems and \(p_i\) represents the associated probability, which satisfies the normalization condition $\sum_{i=0}^{m-1}\ p_i = 1$. Typically, quantum-inspired models leverage a density matrix to represent modality-specific information while capturing the statistical correlations and uncertainties across modalities. As highlighted in \cite{liu2023quantum}, the density matrix formalism is effective for constructing modality representations and modeling cross-modal dependencies. However, these models typically assume that all modalities reside within the same Hilbert space, which restricts the extraction of non-classical correlations.

\subsection{Quantum Jump Method} \label{Sec2:QJ}
The QJ method \cite{plenio1998quantum}, also known as the Monte Carlo wave function method, is a powerful computational tool used to simulate the statistical evolution described by the master equation, but it operates directly on the wave function level, bypassing the density matrix formulation. This method tracks individual quantum trajectories, with each trajectory involving random “jump” events that model irreversible processes such as decoherence and dissipation, driven by system-environment interactions. Unlike the pure unitary evolution, these jumps introduce stochasticity, reflecting the irreversible nature of the interaction between the system and its environment. The transition between states is probabilistic. Specifically, a transition occurs if a randomly generated number $r$ is smaller than the product of $\gamma_{\text{total}}$ and $dt$; otherwise, the system evolves unitarily. The total rate $\gamma_{\text{total}}$ for a pure state $|\psi\rangle$ is given by
\begin{equation}
\gamma_{\text{total}} = \sum_k \gamma_k |\langle \psi | L_k | \psi \rangle|^2,
\end{equation}
where $\gamma_k$ is the jump rate associated with the $k$-th Lindblad operator $L_k$. The Lindblad operators \cite{manzano2020short} describe the dissipative channels responsible for the quantum system's coupling with its environment. Each operator captures a distinct environmental interaction, allowing the model to account for the loss of coherence and other non-unitary effects.
When a jump occurs, the system’s state evolves according to
\begin{equation}
|\psi^{\prime}\rangle = \frac{L_k |\psi\rangle}{\sqrt{\gamma_k}},
\end{equation}
while unitary evolution is governed by
\begin{equation}
|\psi^{\prime}\rangle = e^{\frac{i}{\hbar} H dt} |\psi\rangle,
\end{equation}
where $H$ is the Hamiltonian, $\hbar$ is the reduced Planck's constant, and $dt$ is the time evolution step. This framework allows for both coherent evolution and dissipative dynamics, controlled by the Lindblad operators. 

\subsection{Quantum Measurement}
Quantum measurement can be defined by a set of orthogonal basis states $\{|i\rangle \}_{i=1}^{M}$, known as quantum projections. Prior to measurement, the quantum system exists in a superposition, embodying all possible outcomes simultaneously—reflecting the inherent uncertainty of quantum systems. Upon measurement, the system collapses to a definitive state corresponding to a single outcome with a specific probability, resolving the prior uncertainty. For instance, when measuring a pure state $\arrowvert\psi\rangle = \sum_{i = 0}^{n-1} \alpha_{i} \arrowvert i\rangle$ by projecting onto the measurement basis $\{\arrowvert i\rangle\}$, the quantum state collapses to one of the basis states with probability
\begin{equation}\label{qmdef}
p_{i}(\arrowvert \psi\rangle) = \arrowvert \alpha_{i} \arrowvert^2 =  |\langle i\arrowvert \psi\rangle|^2.
\end{equation}
In a more general context, projective measurements can be performed using any state vector $\arrowvert x \rangle$ (i.e., not just the computational basis), with the probability of obtaining $|x\rangle$ given by
\begin{equation}
p_{x}(\arrowvert \psi\rangle) =|\langle x\arrowvert \psi\rangle|^2.
\label{Measure}
\end{equation}
This allows the use of trainable projections, enabling optimized measurement bases.

\subsection{Related Work}
\subsubsection{Classical Multimodal Fusion Models}
Early multimodal representations were often constructed using the tensor product of unimodal embeddings \cite{zadeh2017tensorFusion}. To address computational complexity, low-rank fusion techniques were introduced \cite{Liu2018EfficientModality}. However, these models failed to capture the inherent continuity and dependencies in unimodal data, prompting the use of recurrent neural networks for contextual modelling \cite{majumder2018multimodal}. The adoption of Transformer-based fusion models \cite{Tsai2019TransformerUnaligned, yin2024token} and cross-modal attention \cite{zhang2023learningAdaptive, Sun2024EfficientDual} further advanced multimodal integration. Recent multitask learning frameworks, such as Self-MM \cite{yu2021learning}, UniMSE \cite{Hu2022UniMSETowards}, and MMML \cite{Wu2024Multimodalloss}, optimize multiple multimodal and unimodal tasks to improve consistency and task differentiation. Besides, metric learning \cite{sun2025multimodal, wang2025sslmm}, graph representation learning \cite{hu2024graph}, transfer learning \cite{wang2025mdkat}, and modal enhancement methods \cite{wu2025enriching, wang2025dlf}, further enhance feature embeddings in multimodal representations. Additionally, the deployment of large pretrained vision-language models, including GPT-4V \cite{lian2024gpt} and PAMoE-MSA \cite{huang2025pamoe}, has advanced multimodal reasoning capabilities in sentiment analysis. However, classical approaches remain limited in interpretability, lacking both model transparency to clarify the forward process and post-hoc mechanisms to extract meaningful insights. Moreover, they are inherently constrained in capturing non-classical correlations across modalities—subtle, inseparable dependencies that often underlie the richness of human language and vision.

\begin{figure*}
    \centering
    \includegraphics[width=1\linewidth]{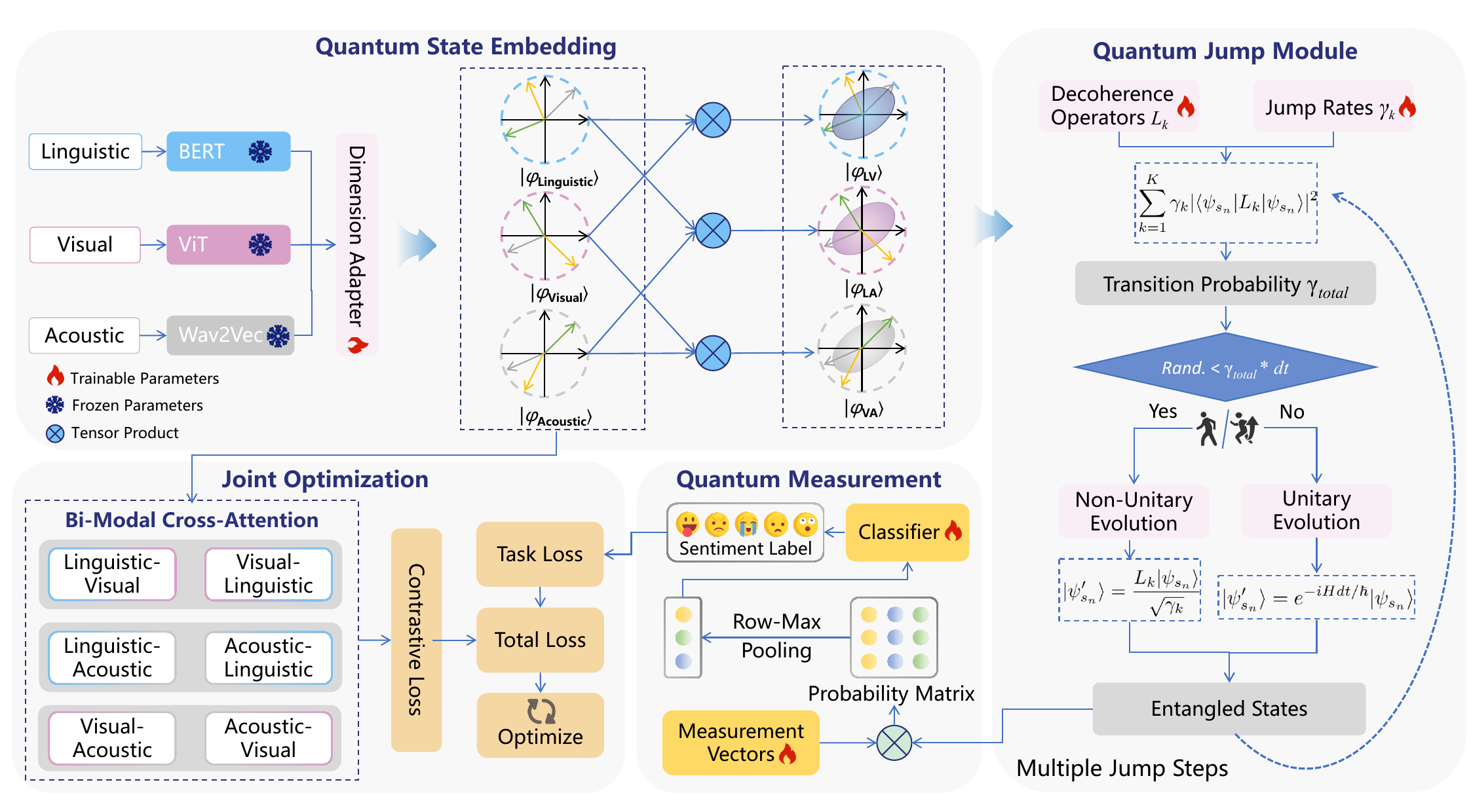}
    \caption{Pipeline of QiNN-QJ. The blue $\otimes$ denotes the tensor product, while the gray $\otimes$ indicates the inner product between vectors. Trainable and non-trainable modules are explicitly distinguished using designated symbols.}
    \label{fig:pipline_model}
\end{figure*}

\subsubsection{Quantum-Inspired Multimodal Models}
Quantum-theoretic principles have recently inspired a new class of machine learning models that address key limitations of classical approaches, particularly in terms of interpretability and modelling complex correlations \cite{gkoumas2018investigating}. These quantum-inspired methods draw on foundational concepts such as superposition, entanglement and interference, which are implemented either through classical neural architectures \cite{liu2023quantum} or quantum circuits \cite{zheng2024quantumCircuits}. For example, QMF \cite{li2021quantumfusionvideo} models intra- and intermodal relationships via superposed and entangled word embeddings, improving the expressive capacity in video sentiment analysis. Similarly, QMSA \cite{ZHANG201821qmsa} introduces density matrix-based representations that capture richer probabilistic and relational structures than conventional vector-based encodings. By leveraging interference effects, QMSA further enhances decision fusion across modalities, leading to more nuanced emotional inference. These methods have demonstrated effectiveness across various applications, including text classification \cite{gao2025qsim}, information retrieval \cite{wu2021natural}, and multimodal inter-variable modelling \cite{tiwari2024quantumfuzzy}. Moreover, their integration with deep learning paradigms, such as graph neural networks \cite{jia2025hierarchical} and attention mechanisms \cite{Gkoumas2021AnEntanglement}, has yielded improved performance and interpretability in multimodal sentiment tasks. Moreover, the fusion strategy introduced in \cite{yan2025quantum} simulates interactions between the fused quantum system and the semantic environment, enabling dynamic and context-aware feature fusion based on density matrix representation. Despite notable advances in quantum-inspired multimodal learning, many existing models are still grounded in density matrix formulations based on classical probabilistic frameworks. These models typically represent multimodal interactions as probabilistic mixtures of independent modality-specific states, which limits their ability to capture complex and inseparable correlations between modalities. More critically, while unitary operators are theoretically globally accessible and capable of generating multimodal entanglement over a large space, they lack inherent mechanisms to control the strength of entanglement. This limitation makes it difficult to achieve stable convergence during training and can lead to poor generalization in real-world applications.

\section{Method} \label{Sec3:Method}

\subsection{Overall Framework}
Figure~\ref{fig:pipline_model} illustrates the overall architecture of the proposed framework, which consists of the following key components.

\begin{itemize}
 \item Quantum State Embedding: The input modalities, such as text, visual, and acoustic, are first encoded using conventional feature extractors. Each modality-specific representation is then normalized into a quantum pure state, defined as a unit-norm complex-valued vector. These states are combined via tensor product to construct an initial separable multimodal quantum state.
\item Quantum Jump Module: The separable quantum state, representing multimodal information, undergoes stochastic dissipation evolution through the QJ method. This process combines deterministic unitary dynamics, governed by a learnable Hamiltonian, with non-unitary stochastic quantum jumps, modelled by learnable Lindblad operators. The resulting entangled state captures context-sensitive and inseparable cross-modal dependencies, which offers a controllable mechanism for modelling complex multimodal interactions.
 \item Quantum Measurement: The generated entangled state is projected onto a set of trainable quantum measurement vectors to extract informative features. The resulting measurement outcomes, interpreted as semantic alignment probabilities, are subsequently passed through a linear classifier to predict sentiment labels.
 \item Joint Optimization: The network is optimized end-to-end using a composite loss that integrates a task-specific classification loss with a contrastive loss. The contrastive component is derived from cross-attention among modality-specific pure states, promoting semantic consistency and discriminative alignment across modalities.
\end{itemize}
The details of each module are explained in the following.

\subsection{Quantum State Embedding}
We leverage state-of-the-art pre-trained encoders, BERT \cite{devlin2018bert} for textual data, ViT \cite{dosovitskiy2020image} for visual content, and Wav2Vec \cite{ravanelli2021speechbrain} for acoustic segments, all of which are trained on large-scale datasets in their respective domains and are well-suited for handling diverse modalities. To enhance training efficiency, we freeze the weights of these encoders and introduce a trainable \textit{Dimension Adapter} module with shared parameters. This module comprises linear transformations, activation functions, and normalization layers, enabling the dimension reduction of feature vectors. In this way, we adapt the extracted features to task-specific requirements without modifying the backbone structures of the pre-trained encoders.

The low-dimensional feature vector of each modality is normalized to form a quantum pure state within its corresponding Hilbert space, denoted as a state vector $|\psi\rangle = [\alpha_{0},\, \cdots,\, \alpha_{i},\, \cdots]^T$, where the normalization condition $\sum_i |\alpha_i|^2 = 1$ ensures unit norm. To represent joint semantic information across multiple modalities, a composite quantum state is defined in the tensor-product Hilbert space $\mathcal{H}_s := \bigotimes_i \mathcal{H}_w^{(i)}$. The resulting joint pure state is expressed as:
\begin{equation}
\arrowvert \psi_s \rangle =  \sum_{i_1,\dots,i_N} \beta_{i_1\dots i_N} \arrowvert i_1 \rangle \otimes \cdots \otimes \arrowvert i_N \rangle,
\label{General_state_joint}
\end{equation}
where the coefficients satisfy the normalization constraint $\sum_{i_1, \dots, i_N} |\beta_{i_1 \dots i_N}|^2 = 1$. The joint state $|\psi_s\rangle$ encapsulates the probability amplitude distribution over the composite semantic space, thereby modelling the entangled interpretation across modalities. Notably, if each modality is represented in a Hilbert space of dimension $D$, then the full joint state $|\psi_s\rangle$ lies in a space of dimensionality $D^N$, where $N$ is the number of modalities involved.

\subsection{Quantum Jump Module} 

The QJ module is inspired by the QJ method introduced in Sec.~\ref{Sec2:QJ}, which models open quantum systems evolving under both unitary dynamics and non-unitary effects induced by environmental interactions. 
Unlike unitary evolution, which lacks control over entanglement strength, the dissipative nature of the QJ method enables precise modulation of entanglement, offering enhanced stability and robustness for multimodal fusion tasks. 
The process begins by constructing modality pairs, yielding $N$ pairs of $N$ modalities. Each pair is encoded as a separable pure state, expressed as the tensor product of the corresponding unimodal quantum states
\begin{equation}
|\psi_{s_n}\rangle = |\psi_n\rangle \otimes |\psi_{n+1}\rangle,
\end{equation}
with $n$ being the index of pairs and each $|\psi_n\rangle$ residing in a Hilbert space. All Hilbert spaces are assumed isomorphic and share a common orthonormal basis, ensuring that the pairwise states $|\psi_{s_n}\rangle$ lie in a well-defined tensor-product space prior to entanglement evolution.

The evolution proceeds through two processes. The first is a unitary evolution governed by a learnable Hamiltonian $H$, capturing coherent system dynamics:
\begin{equation}
|\psi_{s_n}'\rangle = e^{-i H dt / \hbar} |\psi_{s_n}\rangle,
\end{equation}
yielding the intermediate state $|\psi_{s_n}'\rangle$. The second process introduces stochastic environmental interactions via non-unitary evolution. A quantum jump occurs with probability
\begin{equation}
\gamma_{total} = \sum_{k=1}^{K}p_k = \sum_{k=1}^{K}\gamma_k |\langle \psi_{s_n} | L_k | \psi_{s_n} \rangle|^2,
\end{equation}
where $L_k$ is the learnable Lindblad operator corresponding to the $k$-th dissipation channel, $\gamma_k$ is its associated decay rate, and $\gamma_{total}$ is the total jump probability. When a jump occurs, one of the operators $L_k$ is selected at random according to the distribution defined by $\{p_k\}$, and the system collapses to the new state
\begin{equation}
|\psi_{s_n}'\rangle = \frac{L_k |\psi_{s_n}\rangle}{\sqrt{\gamma_k}}.
\end{equation}

To enable differentiable learning, we decouple the stochastic sampling process from the deterministic forward propagation of the network. Specifically, a random variable uniformly sampled from $[0,1]$ determines the evolution: if the sampled value is smaller than $\gamma_{total} * dt$ with $dt$ as the evolution time, a quantum jump occurs. Otherwise, the state undergoes a unitary transformation.
This evolution is repeated a number of times chosen according to the dataset characteristics, allowing the progressive and controllable generation of quantum entanglement across modalities. By jointly modelling coherent and dissipative processes, the QJ module offers a principled and expressive framework for learning entangled multimodal representations.

\subsection{Quantum Measurement} 
The projective measurement operation in quantum mechanics can be interpreted as computing the fidelity between quantum states. Specifically, it quantifies the similarity between the entangled semantic state and a measurement vector, serving as a comparison basis in the Hilbert space. After the evolution process via the QJ module, a set of parametrised measurement vectors $\{ |m_i\rangle \}$ is applied to the entangled state $|\psi_s\rangle$, yielding measurement probability matrix with each entry defined as:
\begin{equation}
q_{in}(|\psi_{s_n}\rangle) = |\langle m_i | \psi_{s_n} \rangle|^2.
\end{equation}
Here, $q_{in}$ denotes the probability that the quantum state $|\psi_{s_n}\rangle$ aligns with the semantic interpretation encoded by the measurement vector $|m_{i}\rangle$. Note that the measurement vectors $\{ |m_i\rangle \}$ are learned in a data-driven fashion. Compared to previous quantum-inspired models that rely on density matrix projections \cite{li2021quantumfusionvideo, yan2025quantum}, our use of pure-state projectors significantly reduces computational complexity from $O(n^3)$ to $O(n)$, enabling more efficient virtual measurement of high-dimensional quantum states.

\subsection{Joint Optimization}

\begin{table*}[]
\centering
\caption{Statistics of Benchmark Datasets.}
\label{tab:statistics-dataset}
\begin{tabular}{@{}ccccccccccc@{}}
\toprule
\multirow{2}{*}{Dataset} & \multicolumn{3}{c}{Train} & \multicolumn{3}{c}{Validation} & \multicolumn{3}{c}{Test} & \multirow{2}{*}{All} \\ \cmidrule(lr){2-10}
 & negative & neutral & positive & negative & neutral & positive & negative & neutral & positive &  \\ \midrule
CMU-MOSI & 522 & 53 & 679 & 92 & 13 & 124 & 379 & 30 & 277 & 2199 \\
CMU-MOSEI & 4738 & 3540 & 8048 & 506 & 433 & 932 & 1350 & 1025 & 2284 & 23453 \\
CH-SIMS & 742 & 207 & 419 & 248 & 69 & 139 & 248 & 69 & 140 & 2281 \\ \bottomrule
\end{tabular}%
\end{table*}

Following the quantum measurement, a row-wise max-pooling operation is applied to the resulting feature matrix, whose entries are denoted by $q_{in}$. The pooled features are then passed through several linear layers to produce the output vector $\hat{y}$. To align predictions with semantic labels, a vector-based similarity metric is incorporated into the task-specific loss, measuring the distance between the predicted feature vector and the ground-truth sentiment label embedding. For a training set of $M$ samples across $C$ sentiment categories, the task loss is defined as
\begin{equation}
\label{eq:task_loss}
\mathcal{L_\mathrm{task}}=-\frac{1}{M}\sum_{m=1}^M\sum_{c=1}^C y^{(m)}_{c}\log(\hat{y}^{(m)}_{c}),
\end{equation}
where \(y_c\) and \(\hat{y}_c\) are the ground-truth label and predicted label for the $c$-th category, respectively.

In addition, the pure state representation of each modality, analogous to a normalized quantum state vector, serves as the foundation for computing pairwise cross-attention. Following the approach in \cite{gao2021simcse}, we introduce a contrastive loss to enhance the discriminative alignment between modalities, formulated as:
\begin{equation}
\label{eq:ct_loss}
\mathcal{L_\mathrm{con}}=-\frac{1}{M}\sum_{m=1}^M\log\frac{\exp(\mathrm{sim}(z_a^{(m)},z_b^{(m)})/\tau)}{\sum_{k=1}^M\exp(\mathrm{sim}(z_a^{(m)},z_b^{(k)})/\tau)},
\end{equation}
where $z_a$ and $z_b$ are cross-attention vectors derived from the pure state representations of two modalities, \(\mathrm{sim}\) denotes the cosine similarity between vectors, \(\tau\) is the temperature hyperparameter and $M$ is the number of training samples. This formulation can be interpreted as encouraging high fidelity between aligned modal states while suppressing similarity to mismatched states, thereby reinforcing robust quantum-inspired inter-modal correlations.

Finally, the total loss is obtained by combining the contrastive loss and task-specific loss. The weighted total loss is expressed as:
\begin{equation}
\label{eq:loss_total}
\mathcal{L}_{\mathrm{total}} = \alpha \mathcal{L}_{\mathrm{task}} + \beta \mathcal{L}_{\mathrm{con}}
\end{equation}
where $\alpha$ and $\beta$ are trainable weights that balance the contributions of each loss term.

\section{Experiments} \label{Sec4:Results}
\subsection{Experiment Details}
\subsubsection{Datasets}
The experiments were conducted on three benchmark datasets for multimodal sentiment analysis, including CMU-MOSI \cite{zadeh2016mosi}, CMU-MOSEI \cite{zadeh2018multimodal} and CH-SIMS \cite{yu2020chsims}. Specifically, the CMU-MOSI dataset comprises 2,199 labeled YouTube video clips sourced from 89 distinct speakers. Each clip integrates three modalities—text, audio, and visual—and is annotated with sentiment intensity on a continuous scale ranging from -3 (strongly negative) to +3 (strongly positive). As an extension of CMU-MOSI, the CMU-MOSEI dataset significantly expands the scope, featuring 23,453 sentence-level video clips. These clips encompass 1,000 speakers discussing 250 different topics. For Chinese-language contexts, the CH-SIMS dataset offers a specialized resource, comprising 2,281 meticulously curated movie clips that depict real-life scenarios. A detailed summary of the dataset statistics is presented in Table~\ref{tab:statistics-dataset}.

\subsubsection{Evaluation metrics}
In alignment with prior studies \cite{zhang2023learningAdaptive}, we adopt a set of standard evaluation metrics. Specifically, on the CMU-MOSI and CMU-MOSEI datasets, performance is assessed using seven-class accuracy (denoted by Acc-7), binary accuracy (denoted by Acc-2), F1 score, Mean Absolute Error (MAE), and correlation coefficient (denoted by Corr). For the CH-SIMS dataset, consistent with the protocol with previous works, we report results in terms of five-class accuracy (denoted by Acc-5), three-class accuracy (denoted by Acc-3), and Acc-2. Notably, for all metrics except MAE, higher values signify superior performance, whereas lower MAE values are indicative of better model accuracy. 

\begin{table}[]
\centering
\caption{Detailed configurations for QiNN-QJ on benchmark datasets.}
\label{tab:param-configurations}
\resizebox{\columnwidth}{!}{%
\begin{tabular}{@{}clccc@{}}
\toprule
 &  & \multicolumn{3}{c}{} \\
 &  & \multicolumn{3}{c}{\multirow{-2}{*}{\textbf{Dataset}}} \\ \cmidrule(l){3-5} 
 &  &  &  &  \\
\multirow{-4}{*}{\textbf{Category}} & \multirow{-4}{*}{\textbf{Parameter}} & \multirow{-2}{*}{\textbf{CMU-MOSI}} & \multirow{-2}{*}{\textbf{CMU-MOSEI}} & \multirow{-2}{*}{\textbf{CH-SIMS}} \\ \midrule
\multirow{-2}{*}{\textbf{Preprocessing}} & \begin{tabular}[c]{@{}l@{}}Feature Dimension \\ (Input)\end{tabular} & \begin{tabular}[c]{@{}c@{}}T:768, V:5, \\A:20\end{tabular} & \begin{tabular}[c]{@{}c@{}}T:768, V:35,\\ A:74\end{tabular} & \begin{tabular}[c]{@{}c@{}}T:768, V:709,\\ A:33\end{tabular} \\ \midrule
 & Encoder Type & BERT & BERT & BERT \\
& Output Dimension & 768 & 768 & 768 \\
\multirow{-3}{*}{\textbf{\begin{tabular}[c]{@{}c@{}}Text \\ Encoder\end{tabular}}} & Adapter Dimension & 10 & 10 & 10 \\ \midrule
 & Encoder Type & Wav2Vec 2.0 & Wav2Vec 2.0 & Wav2Vec 2.0\\
 & {\color[HTML]{333333} Number of Layers} & 12 & 12 & 12 \\
 & Output Dimension & 768 & 768 & 768 \\ 
\multirow{-4}{*}{\textbf{\begin{tabular}[c]{@{}c@{}}Audio \\ Encoder\end{tabular}}} & Adapter Dimension & 10 & 10 & 10 \\ \midrule
 & Encoder Type & ViT& ViT & ViT  \\
 & Output Dimension & 768 & 768 & 768 \\ 
\multirow{-3}{*}{\textbf{\begin{tabular}[c]{@{}c@{}}Vision \\ Encoder\end{tabular}}} & Adapter Dimension & 10 & 10 & 10 \\ \midrule
 & Number of Time Step & 20 & 10 & 20 \\
\multirow{-2}{*}{\textbf{\begin{tabular}[c]{@{}c@{}}Quantum \\ Jump Module \end{tabular}}} & \begin{tabular}[c]{@{}l@{}}Hamitonian/Operator Dimension\end{tabular} & 100 &100 &100 \\ \midrule
 & Similarity Funtion & cosine & cosine & cosine \\
\multirow{-2}{*}{\textbf{\begin{tabular}[c]{@{}c@{}}Contrastive \\ Learning\end{tabular}}} & Normalized Temperature & 0.07 & 0.07 & 0.05 \\ \midrule
 & Batch Size & 32 & 16 & 32 \\
 & Optimizer & AdamW & AdamW & AdamW \\
 & Learning Rate Text & 5e-5 & 5e-6 & 5e-6 \\
 & Learning Rate Audio & 2e-3 & 1e-4 & 1e-3 \\
 & Learning Rate Video & 2e-4 & 2e-5 & 5e-5 \\
 & Weight Decay & 5e-3 & 5e-3 & 5e-3 \\
 & Dropout & 0.1 & 0.15 & 0.1 \\
 & Training Epochs & 50 & 50 & 50 \\
\multirow{-9}{*}{\textbf{Optimization}} & Early Stopping Epochs & 5 & 5 & 5 \\ \bottomrule
\end{tabular}%
}
\begin{tablenotes}
    \footnotesize
    \item \textit{T, V and A in Feature\_dims denote text, video and audio, respectively.}
\end{tablenotes}
\end{table}

\begin{table*}[]
\centering
\caption{Experimental results of comparison models on the CMU-MOSI and CMU-MOSEI datasets.}
\label{tab:exp_res_cmu}
\begin{tabular}{@{}ccccccc|ccccc@{}}
\toprule
\multirow{2}{*}{Category} & \multirow{2}{*}{Method} & \multicolumn{5}{c|}{CMU-MOSI} & \multicolumn{5}{c}{CMU-MOSEI} \\ \cmidrule(l){3-12} 
 &  & Acc-2 & F1-Score & Acc-7 & MAE & Corr & Acc-2 & F1-Score & Acc-7 & MAE & Corr \\ \midrule
\multirow{7}{*}{Classical} & UniMSE (2022) \cite{Hu2022UniMSETowards} & 86.9 & 86.42 & 48.68 & 0.691 & 0.809 & 87.5 & 87.46 & 54.39 & 0.523 & 0.773 \\
 & ALMT (2023) \cite{zhang2023learningAdaptive} & 86.43 & 86.47 & 49.42 & 0.683 & 0.805 & 86.79 & 86.86 & 54.28 & 0.526 & 0.779 \\
 & MMML (2024) \cite{Wu2024Multimodalloss} & \textbf{88.16} & \textbf{88.15} & 48.25 & 0.642 & 0.838 & 86.73 & 86.49 & {\ul 54.95} & 0.517 & 0.79 \\
 & GRAFN (2025) \cite{hu2024graph} & 86.06 & 85.94 & 50.6 & \textbf{0.601} &  {\ul 0.864} & 84.65 & 84.68 & 53.4 & 0.634 & 0.79 \\
 & DLF (2025) \cite{wang2025dlf} & 85.06 & 85.04 & 47.08 & 0.731 & 0.781 & 85.42 & 85.27 & 53.9 & 0.536 & 0.764 \\
 & MDKAT (2025) \cite{wang2025mdkat} & 85.6 & 85.6 & 46.2 & 0.717 & / & 86.5 & 86.4 & 54.3 & 0.532 & / \\
 & MIMRL (2025) \cite{sun2025multimodal} & 86.9 & 86.9 & 47.1 & 0.687 & 0.792 & 86.4 & 86.4 & 55.4 & \textbf{0.513} & 0.801 \\ \midrule
\multirow{4}{*}{Quantum-inspired} & QMF (2021) \cite{li2021quantumfusionvideo} & 79.74 & 79.62 & 33.53 & 0.914 & 0.695 & 80.69 & 79.77 & 47.88 & 0.639 & 0.657 \\
 & EFNN (2021) \cite{Gkoumas2021AnEntanglement} & 80.9 & 80.8 & 35.9 & 0.91 & 0.69 & 82.8 & 82.6 & 50.2 & 0.6 & 0.69 \\ \cmidrule(l){2-12} 
 & QiNN-QJ & {\ul 87.59} & {\ul 87.62} & \textbf{54.52} & {\ul 0.613} & \textbf{0.88} & \textbf{88.2} & \textbf{88.18} & \textbf{55.63} & {\ul 0.515} & \textbf{0.811} \\ \bottomrule
\end{tabular}%
\begin{tablenotes}
    \footnotesize
    \item \textit{The best results are highlighted in \textbf{bold}. The second-best results are highlighted with an {\ul underline}.}
\end{tablenotes}
\end{table*}

\begin{table*}[]
\centering
\caption{Experimental results of comparison models on the CH-SIMS dataset.}
\label{tab:exp_res_sims}
\begin{tabular}{@{}ccccccccc@{}}
\toprule
\multirow{2}{*}{Category} &\multirow{2}{*}{Method} & \multicolumn{6}{c}{CH-SIMS} \\ \cmidrule(l){3-8} 
\multicolumn{1}{c}{}& & Acc-2 & F1-Score & Acc-3 & Acc-5 & MAE & Corr \\ \midrule
\multirow{9}{*}{Classical} &TFN (2017) \cite{zadeh2017tensorFusion} & 78.38 & 78.62 & 65.12 & 39.3 & 0.432 & 0.591 \\
&LMF (2018) \cite{Liu2018EfficientModality} & 77.77 & 77.88 & 64.68 & 40.53 & 0.441 & 0.576 \\
 &MulT (2019) \cite{Tsai2019TransformerUnaligned} & 78.56 & 79.66 & 64.77 & 37.94 & 0.453 & 0.564 \\
 &Self-MM (2021) \cite{yu2021learning} & 80.04 & 80.44 & 65.47 & 41.53 & 0.425 & 0.595 \\
 &ALMT (2023) \cite{zhang2023learningAdaptive} & 81.19 & {\ul 81.57} & {\ul 68.93} & 45.73 & 0.404 & 0.619 \\
 &EMT (2024) \cite{Sun2024EfficientDual} & 80.1 & 80.1 & 67.4 & 43.5 & {\ul 0.396} & 0.623 \\
 &TMT (2024) \cite{yin2024token} & 80.53 & 81.11 & 68.71 & {\ul 45.73} & / & / \\
 &SSLMM (2025) \cite{wang2025sslmm} & 80.06 & 80.56 & 65.48 & 39.52 & 0.409 & 0.589 \\
  &DEVA (2025) \cite{wu2025enriching} & 79.64 & 80.32 & 65.42 & 43.07 & 0.424 & 0.583 \\ \midrule
\multirow{2}{*}{LLM} &GPT-4V (2024) \cite{lian2024gpt} & 81.24 & / & / & / & / & / \\
 &PAMoE-MSA (2025) \cite{huang2025pamoe} & {\ul 81.6} & 81.5 & / & / & 0.41 & {\ul 0.633} \\\midrule
Quantum-inspired &QiNN-QJ & \textbf{81.87} & \textbf{82.04} & \textbf{69.83} & \textbf{46.92} & \textbf{0.381} & \textbf{0.689} \\ \bottomrule
\end{tabular}%
\begin{tablenotes}
    \footnotesize
    \item \textit{The best results are highlighted in \textbf{bold}. The second-best results are highlighted with an {\ul underline}.}
\end{tablenotes}
\end{table*}

\subsubsection{Baselines}
To comprehensively assess the proposed model, we have selected a diverse array of baseline models for comparison. These models are systematically classified based on their underlying architectural principles and technological evolution paths. The specifics are elaborated below.
\begin{itemize}
\item Foundation Fusion Methods: Early studies have primarily concentrated on developing fusion architectures that directly incorporate cross-modal features while simultaneously capturing intra-modal and inter-modal dynamics. Among the notable models in this domain are the Tensor Fusion Network (TFN) \cite{zadeh2017tensorFusion}, Low-rank Multimodal Fusion (LMF) \cite{Liu2018EfficientModality}, and Multimodal Transformer (MulT) \cite{Tsai2019TransformerUnaligned}.
\item Guided Learning-Based Fusion Methods: These models enhance learning efficiency and generalization by incorporating strategies like multi-task learning, self-supervised learning, or robust modality guidance. Examples include Self-Supervised Multi-Task Learning (Self-MM) \cite{yu2021learning}, Unified Multimodal Sentiment Analysis and Emotion Recognition (UniMSE) \cite{Hu2022UniMSETowards}, Adaptive Language-guided Multimodal Transformer (ALMT) \cite{zhang2023learningAdaptive}, and Multimodal Multi-loss Fusion Network (MMML) \cite{Wu2024Multimodalloss}, which optimize cross-modal interactions using diverse loss functions.
\item Represent Learning-Based Methods: These methods focus on creating low-dimensional vector representations to capture emotion-related features while enhancing cross-modal information consistency. Notable approaches include Efficient Multimodal Transformer (EMT) \cite{Sun2024EfficientDual}, Token-disentangling Mutual Transformer (TMT) \cite{yin2024token}, Semi-Supervised Learning with Missing Modalities (SSLMM) \cite{wang2025sslmm}, DEVA \cite{wu2025enriching}, GRAFN \cite{hu2024graph}, DLF \cite{wang2025dlf}, MDKAT \cite{wang2025mdkat}, and MIMRL \cite{sun2025multimodal}.
\item Large Language Models (LLMs): We adopt advanced methods based on state-of-the-art multimodal large language models and related technologies, including GPT-4V \cite{lian2024gpt} and the Polarity-aware Mixture of Experts (PAMoE-MSA) \cite{huang2025pamoe}.
\item Quantum-Inspired Methods: These methods leverage the unique properties of quantum theory to address the limitations encountered by classical systems when processing complex multimodal data. Specifically, we highlight two notable approaches that have been successfully applied to such datasets: QMF \cite{li2021quantumfusionvideo}, and the Entanglement-Driven Fusion Neural Network (EFNN) \cite{Gkoumas2021AnEntanglement}.
\end{itemize}

\subsubsection{Implementation Details}
We implemented the proposed model in PyTorch (Python 3.11.4) and conducted experiments on an NVIDIA RTX 4090 GPU with 64 GB RAM. Hyperparameters were optimized via an extensive grid search over predefined ranges. In particular, the adapter dimension, which controls the dimensionality of pure states, was selected from $\{6, 8, 10, 12, 14\}$. The effect of varying the number of time steps is analysed in Fig.~\ref{fig:poshoc}. The model was executed five times under distinct random seeds to ensure robustness and reliability. Through comprehensive evaluation, the set demonstrating the most favourable overall performance was selected as the definitive outcome. The detailed configurations are presented in Table~\ref{tab:param-configurations}.

\subsection{Performance}

\subsubsection{CMU-MOS(E)I Datasets}

Table~\ref{tab:exp_res_cmu} presents the experimental results on the CMU-MOSI and CMU-MOSEI datasets, showing that the proposed model achieves optimal or near-optimal performance across all evaluation metrics. On CMU-MOSI, QiNN-QJ ranks second only to NMML in Acc-2 and F1-Score, and to DLF in MAE, surpassing all other state-of-the-art classical models. Notably, it delivers a substantial improvement on the Acc-7 metric, which reflects the model’s fine-grained sentiment discrimination capability that is critical for nuanced sentiment understanding. On CMU-MOSEI, QiNN-QJ is outperformed only by MIMRL in MAE, while exceeding all other methods on the remaining metrics, including a clear lead in Acc-7. In particular, the proposed model markedly outperforms all previously applied quantum-inspired methods on these two datasets.

\subsubsection{CH-SIMS}
The previous two datasets contain English multimodal sentiment analysis data, primarily reflecting expression norms and emotional communication patterns in Western cultural contexts. To systematically assess the cross-linguistic and cross-cultural generalizability of the proposed model, we further evaluated it on the Chinese multimodal sentiment analysis dataset CH-SIMS, which covers lexical, syntactic and pragmatic levels along with its culture-specific expressions, differing significantly from English data. This distinction enables an objective evaluation of the model’s adaptability and robustness across languages and cultural settings. 

Experimental results are reported in Table~\ref{tab:exp_res_sims}.
QiNN-QJ consistently outperforms both domain-specific models and multimodal LLMs across all key evaluation metrics. In particular, Acc-5, which measures fine-grained sentiment classification across five levels, directly reflects the ability to capture nuanced emotional distinctions beyond simple polarity. Similarly, the Corr metric provides a robust measure of how closely the model’s predictions align with human-labelled sentiment intensity, reflecting both accuracy and ranking consistency. On both metrics, our method achieves a substantial margin over state-of-the-art methods. Notably, QiNN-QJ is the first quantum-inspired model applied on the multimodal dataset in Chinese.
These results confirm the strong generalization capability of our approach, enabling accurate, fine-grained sentiment analysis across diverse linguistic and cultural contexts, and providing a solid foundation for its future deployment in real-world applications.

\subsection{Ablation Test}
To systematically examine the influence of different modalities and core components on model performance, we conducted ablation studies on the CMU-MOSI and CMU-MOSEI datasets. The experiments assessed the effects of removing individual modality inputs, as well as eliminating key modules from the architecture. As reported in Table~\ref{tab:ablation_exp}, performance drops markedly when either $\mathcal{L_\mathrm{con}}$ or the QJ module is excluded, particularly in the case of QJ, highlighting their pivotal roles in the proposed quantum-inspired framework. Furthermore, the degradation observed in both single-modal and dual-modal settings confirms the importance of multimodal fusion. Among modalities, text consistently contributes the most to predictive accuracy, while audio provides relatively limited gains.

To further evaluate the efficacy of different fusion strategies, Table~\ref{tab:EffectsTechniques} compares their impact on CMU-MOSI and CH-SIMS. The density matrix representation with concatenation encodes modal information and interrelations, but it underperforms due to the complexity and abstraction of density matrix construction, which hinders effective salience control compared to tensor concatenation. In contrast, density matrices with classical probability add dynamically weight modalities, enhancing representational capacity and yielding superior performance across both datasets compared to the concatenation one. The pure state representation through concatenation performs well by explicitly modelling intermodal feature interactions but fails to capture quantum entanglement, limiting its cross-modal expressiveness. In contrast, the pure state representation with Unitary Transformation (UT) performs slightly better due to entanglement modelling, though not in the optimal manner. The proposed model utilizes the QJ method to simulate the dissipative evolution of an open quantum system, effectively modelling entanglement among modalities, thereby achieving the best overall performance.

\begin{table*}[]
\centering
\caption{Ablation results for QiNN-QJ on the CMU-MOSI and CMU-MOSEI datasets.}
\label{tab:ablation_exp}
\begin{tabular}{@{}lccccc|ccccc@{}}
\toprule
\multicolumn{1}{c}{\multirow{2}{*}{Methods}} & \multicolumn{5}{c|}{CMU-MOSI} & \multicolumn{5}{c}{CMU-MOSEI} \\ \cmidrule(l){2-11} 
\multicolumn{1}{c}{} & Acc-2 & F1-Score & Acc-7 & MAE & Corr & Acc-2 & F1-Score & Acc-7 & MAE & Corr \\ \midrule
QiNN (Text Only) & 83.78 & 83.79 & 46.83 & 0.701 & 0.803 & 86.12 & 86.35 & 54.38 & 0.548 & 0.753 \\
QiNN (Audio Only) & 59.41 & 43.96 & 24.84 & 1.396 & 0.045 & 64.12 & 60.94 & 44.66 & 0.745 & 0.163 \\
QiNN (Video Only) & 63.72 & 54.02 & 26.41 & 1.213 & 0.162 & 62.22 & 60.18 & 45.84 & 0.728 & 0.224 \\ \midrule
QiNN-QJ w/o video & 86.12 & 86.11 & 49.45 & 0.712 & 0.832 & 80.11 & 84.16 & 54.73 & 0.527 & 0.744 \\
QiNN-QJ w/o audio & {\ul 86.67} & {\ul 86.52} & 49.11 & 0.676 & {\ul 0.834} & 85.74 & 85.65 & {\ul 54.91} & {\ul 0.526} & 0.756 \\
QiNN-QJ w/o text & 57.92 & 56.84 & 26.32 & 1.396 & 0.112 & 60.17 & 59.32 & 45.69 & 0.792 & 0.203 \\ \midrule
QiNN-QJ w/o QJ & 85.12 & 85.16 & 47.36 & 0.714 & 0.804 & 85.27 & 85.52 & 52.46 & 0.549 & 0.752 \\
QiNN-QJ w/o $\mathcal{L_\mathrm{con}}$ & 86.42 & 86.24 & {\ul 51.29} & {\ul 0.646} & 0.833 & {\ul 86.36} & {\ul 86.43} & 53.73 & 0.53 & {\ul 0.766} \\ \midrule
QiNN-QJ (Full Model) & \textbf{87.59} & \textbf{87.62} & \textbf{54.52} & \textbf{0.613} & \textbf{0.88} & \textbf{88.2} & \textbf{88.18} & \textbf{55.63} & \textbf{0.515} & \textbf{0.811} \\ \bottomrule
\end{tabular}%
\begin{tablenotes}
    \footnotesize
    \item \textit{The best results are highlighted in \textbf{bold}. The second-best results are highlighted with an {\ul underline}.}
\end{tablenotes}
\end{table*}

\begin{table*}[]
\centering
\caption{{Results of different fusion techniques in QiNN models.}}
\label{tab:EffectsTechniques}
\begin{tabular}{@{}cccccc|cccccc@{}}
\toprule
\multirow{2}{*}{Fusion Technique} & \multicolumn{5}{c|}{CMU-MOSI} & \multicolumn{6}{c}{CH-SIMS} \\ \cmidrule(l){2-12} 
 & Acc-2 & F1-Score & Acc-7 & MAE & Corr & Acc-2 & F1-Score & Acc-3 & Acc-5 & MAE & Corr \\ \midrule
Density Matrix + Concatenation 
& 83.89 & 83.86 & 40.71 & 0.732 & 0.789 & 75.51 & 75.52 & 60.27 & 38.62 & 0.427 & 0.583 \\ 
Density Matrix + Addition & 85.31 & 85.30 & 42.41 & 0.698 & 0.826 & 76.29 & 76.36 & 63.08 & 37.84 & 0.431 & 0.584 \\  \midrule
Pure State + Tensor Product & 85.12 & 85.16 & 47.36 & 0.714 & 0.804 & 75.22 & 75.79 & 66.29 & 41.52 & 0.421 & 0.613 \\ 
Pure State + Tensor Product + UT & 85.27 & 85.31 & 46.43 & 0.686 & 0.820 & 75.92 & 75.85 & 65.86 & 41.88 & 0.415 & 0.629 \\ 
Pure State + Tensor Product + QJ& \textbf{87.59} & \textbf{87.62} & \textbf{54.52} & \textbf{0.613} & \textbf{0.880} & \textbf{81.87} & \textbf{82.04} & \textbf{69.83} & \textbf{46.92} & \textbf{0.381} & \textbf{0.689} \\ \bottomrule
\end{tabular}%

\begin{tablenotes}
    \footnotesize
    \item \textit{The best results are highlighted in \textbf{bold}.}
\end{tablenotes}
\end{table*}

\section{Post-Hoc Interpretability and Robustness Analysis} \label{Sec5:Analysis}
\subsection{Entanglement Metric}
To explain the efficacy of QJ in multimodal data fusion, we adopt the von-Neumann entanglement entropy $\mathcal{S}$ \cite{horodecki1994quantum} as a metric to quantify the degree of entanglement among components within a quantum system. It is computed as follows:
\begin{equation}
\label{eq:VonNeumannEntanglement}
    \mathcal{S}=-\sum_{i=1}^K|\lambda_i|^2\mathrm{log}(|\lambda_i|^2),
\end{equation}
where $\lambda_i$ is the Schmidt coefficient of the composite pure state and $K$ is the dimension of the subsystems. In quantum systems, high entropy values suggests the presence of complex non-classical interactions. Conversely, a decline in entropy reflects a transition toward a stable and predictable state.

\begin{figure}[]
    \centering
    \includegraphics[width=1\linewidth]{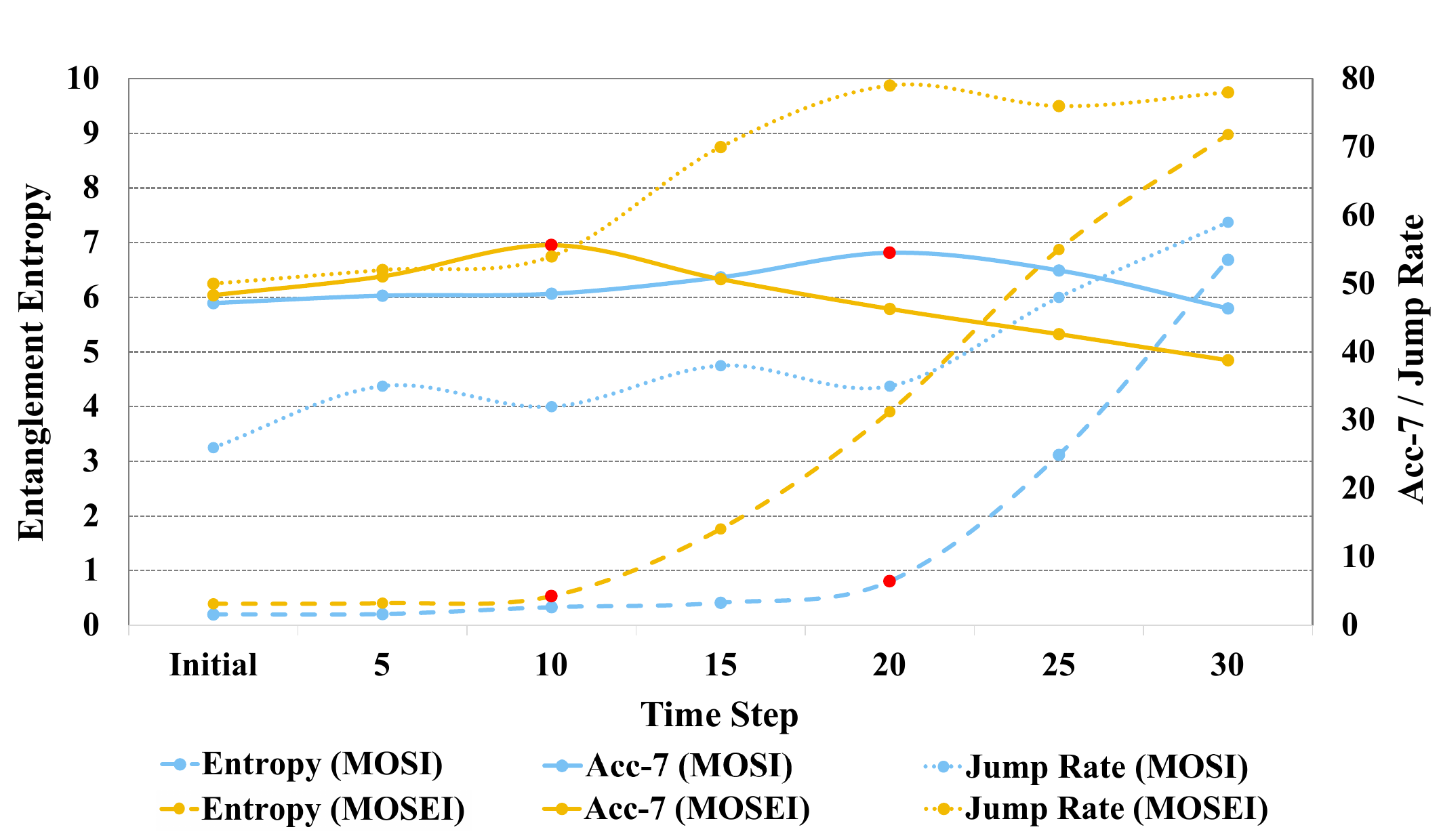}
    \caption{Variation of von-Neumann entanglement entropy, Acc-7 and jump rate across time steps for the text and video modalities.}
    \label{fig:poshoc}
\end{figure}

\begin{figure*}
    \centering
    \includegraphics[width=0.85\linewidth]{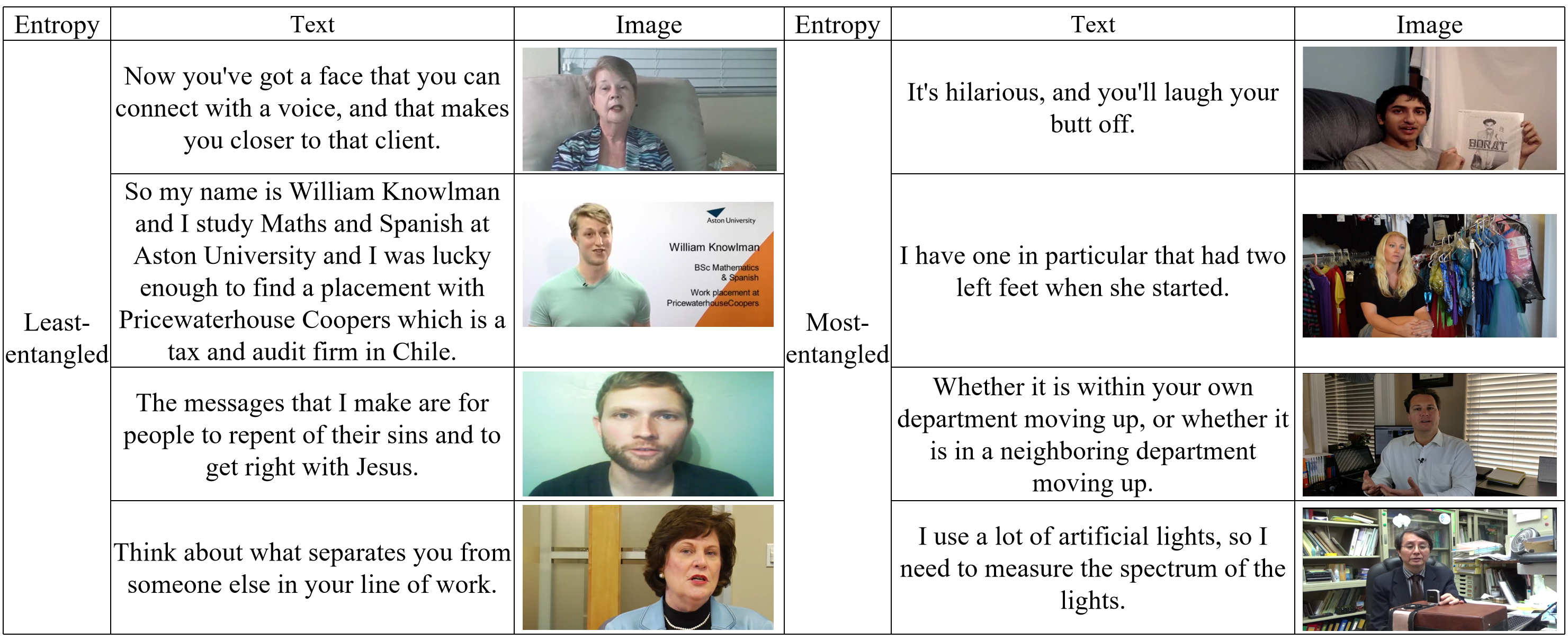}
    \caption{Visualization of Least and Most Entangled Pairs. The left columns present the selected least-entangled pairs, while the right columns show the most-entangled pairs. Each part includes the textual input alongside its corresponding image for comparison.}
    \label{fig:EntangledPairs}
\end{figure*}

\subsection{Controllable Evolution of Entanglement}
Figure~\ref{fig:poshoc} presents the temporal evolution of entanglement entropy, jump rate and accuracy for QiNN-QJ on two datasets. Red dots denote peak accuracy and inflection points in the entanglement entropy curve. We can see that prior to the inflection point, accuracy increases steadily, implying that the model is effectively extracting complex intermodal entanglement. After the inflection point, when entanglement entropy rises, accuracy declines, which suggests overemphasis on intricate correlations at the expense of simpler and informative ones. Simultaneously, the jump rate becomes more volatile, which reflects heightened uncertainty.

Unlike purely unitary transformations, QiNN-QJ incorporates quantum dissipation via learnable Lindblad operators, allowing the stochastic evolution to regulate excessive entanglement growth. This dissipative mechanism introduces structured randomness and steady-state attractors, thereby stabilizing training and preventing uncontrolled correlation escalation. As a result, the entanglement evolution in QiNN-QJ remains controllable, enabling the model to balance between capturing essential cross-modal dependencies and avoiding overfitting. Notably, the inflection points differ across datasets, indicating dataset-specific intermodal correlation structures and the necessity for QiNN-QJ to adaptively determine an optimal termination time for random evolution under dissipative regulation.


\begin{table*}[!htbp]
\centering
\caption{Performance comparison of QiNN-QJ and Self-MM across varying modality-missing rates.}
\label{tab:modality-missing}
\begin{tabular}{@{}cccccccccc@{}}
\toprule
\multirow{2}{*}{Datasets} & \multirow{2}{*}{Methods} & \multirow{2}{*}{Metrics} & \multicolumn{6}{c}{Masking Scope (Text + Audio + Video)} & \multirow{2}{*}{Avg. Drop Rate (\%)} \\ \cmidrule(lr){4-9}
 &  &  & No Masking & 10\% & 20\% & 30\% & 40\% & 50\% &  \\ \midrule
\multirow{6}{*}{CMU-MOSI} & \multirow{3}{*}{QiNN-QJ}
  & F1-Score & 87.62 & 87.56 & 85.05 & 85.61 & 85.08 & 85.01 & 0.596 \\
 &  & Acc-7 & 54.52 & 49.28 & 47.81 & 48.03 & 47.66 & 47.58 & 2.614 \\
 &  & Corr & 0.88 & 0.823 & 0.797 & 0.808 & 0.788 & 0.783 & 2.273 \\ \cmidrule(l){2-10} 
 & \multirow{3}{*}{Self-MM} 
  & F1-Score & 85.95 & 85.19 & 85.19 & 83.65 & 83.08 & 82.47 & 0.822 \\
 &  & Acc-7 & 46.4 & 43.67 & 45.29 & 42.41 & 41.67 & 39.85 & 2.929 \\
 &  & Corr & 0.798 & 0.798 & 0.801 & 0.756 & 0.689 & 0.674 & 3.256 \\ \midrule
\multirow{6}{*}{CMU-MOSEI} & \multirow{3}{*}{QiNN-QJ} 
 & F1-Score & 88.18 & 87.93 & 86.96 & 87.09 & 86.69 & 86.51 & 0.381 \\
 &  & Acc-7 & 55.63 & 54.83 & 54.56 & 54.63 & 54.24 & 54.08 & 0.562 \\
 &  & Corr & 0.811 & 0.782 & 0.741 & 0.788 & 0.735 & 0.737 & 1.786 \\ \cmidrule(l){2-10} 
 & \multirow{3}{*}{Self-MM} 
   & F1-Score & 85.3 & 84.81 & 85.25 & 83.66 & 83.14 & 82.65 & 0.626 \\
 &  & Acc-7 & 53.6 & 52.84 & 53.57 & 52.93 & 52.07 & 50.25 & 1.27 \\
 &  & Corr & 0.765 & 0.742 & 0.741 & 0.729 & 0.713 & 0.691 & 2.008 \\ \bottomrule
\end{tabular}%
\end{table*}

\subsection{Visualization of Most and Least Entangled Pairs}

In this section, we present the analysis of the most and least entangled modality pairs based on the QiNN-QJ framework. Using von Neumann entropy as a measure of entanglement, we highlight the modality pairs that exhibit the strongest and weakest cross-modal dependencies. By examining these pairs, we gain deeper insights into the behaviour of the proposed model, especially in how it learns to represent complex and inseparable correlations across modalities.
As shown in \fig{fig:EntangledPairs}, low-entanglement cases typically correspond to clean visual backgrounds with clearly identifiable subjects and relatively simple textual structures. These scenarios reflect weak intermodal dependencies, where information can be effectively interpreted within each modality independently. In contrast, high-entanglement cases are characterized by visually cluttered or noisy backgrounds and text with frequent use of pronouns or context-dependent expressions. In such instances, the semantic meaning cannot be inferred from a single modality in isolation but instead emerges through complex intermodal interactions. This comparison provides deeper insight into how QiNN-QJ adaptively models inseparable and context-sensitive correlations across modalities.


\subsection{Incomplete Modality Robustness Analysis}
To assess robustness under modality absence, we employed a randomized masking protocol in which a fixed proportion of sequential features from each modality was randomly removed in the test set, simulating varying degrees of missing data. Five absence rates (10–50\%) were examined, with independent trials at each level. Results in Table~\ref{tab:modality-missing} show a monotonic performance degradation with increasing masking, confirming that predictive efficacy declines under multimodal data loss. Across all masking ratios, our method consistently outperforms the prior state-of-the-art Self-MM, with the advantage most pronounced at higher absence levels. Average decline rates reveal that both models are more sensitive in MAE and Corr than in Accuracy and F1-score. These findings indicate that our approach exhibits superior adaptability to incomplete multimodal inputs, underscoring its potential reliability in practical scenarios.

\section{Conclusion} \label{Sec6:Conclusion}
In this work, we proposed QiNN-QJ, a quantum-inspired neural network that incorporates the QJ formalism to convert initially separable multimodal product states into controllable entangled representations. By preserving the expressive capacity of unitary evolution while embedding structured stochasticity and steady-state attractor dynamics, QiNN-QJ enables physically interpretable and controllable modelling of cross-modal entanglement within an end-to-end framework. The proposed architecture was fully implemented on classical hardware yet retains theoretical consistency with open quantum system dynamics.

Comprehensive experiments on benchmark multimodal sentiment datasets demonstrate that QiNN-QJ consistently surpasses state-of-the-art classical and quantum-inspired baselines in predictive accuracy, interpretability, and resilience under incomplete modality scenarios. Quantitative entanglement analysis and t-SNE–based visualizations further substantiate its capacity to capture and disentangle meaningful multimodal correlations. Future research will focus on extending QiNN-QJ to variational quantum circuit implementations, cross-domain multimodal transfer, and uncertainty-aware quantum-inspired reasoning for trustworthy and explainable multimodal fusion.

\bibliographystyle{unsrt}
\bibliography{ref}

\end{document}